\documentclass{article}

\usepackage{microtype}
\usepackage{graphicx}
\usepackage{float}
\usepackage{subfigure}
\usepackage{booktabs} % for professional tables
\usepackage{hyperref}
\usepackage{dblfloatfix}
\usepackage{caption} 
\usepackage{amsmath}
\usepackage{amssymb}
\usepackage{mathtools}
\usepackage{booktabs}
\usepackage{multirow}
\usepackage{makecell}
\usepackage[table]{xcolor}
\definecolor{m2blue}{HTML}{e1f5fe}
\definecolor{ball_speed}{HTML}{EEF4FF}
\definecolor{paddle_length}{HTML}{EDF7ED}
\definecolor{paddle_speed}{HTML}{F2F2F2}

\usepackage{lmodern}

\usepackage{hyperref}

\usepackage[accepted]{icml2026}

\usepackage{amsmath}
\usepackage{amssymb}
\usepackage{mathtools}
\usepackage{amsthm}

\usepackage[capitalize,noabbrev]{cleveref}

\theoremstyle{plain}

\theoremstyle{definition}

\theoremstyle{remark}

\usepackage[textsize=tiny]{todonotes}

\icmltitlerunning{CDRL: Reinforcement Learning Inspired by Cerebellar Circuits and Dendritic Mechanisms}

\begin{document}

\twocolumn[
\icmltitle{CDRL: A Reinforcement Learning Framework Inspired by \\ Cerebellar Circuits and Dendritic Computational Strategies }




  \begin{icmlauthorlist}
    \icmlauthor{Sibo Zhang}{sch}
    \icmlauthor{Rui Jing}{sch}
    \icmlauthor{Liangfu Lv}{sch}
    \icmlauthor{Jian Zhang}{sch}
    \icmlauthor{Yunliang Zang}{sch}
    
  \end{icmlauthorlist}

  % \icmlaffiliation{comp}{Xiamen Intretech Inc, Xiamen, Fujian, China}
  \icmlaffiliation{sch}{Academy of Medical Engineering and Translational Medicine, Tianjin University, Tianjin, China}

  \icmlcorrespondingauthor{Yunliang Zang}{yunliangzang@tju.edu.cn}

  \icmlkeywords{Machine Learning, ICML}

  \vskip 0.3in
]



\printAffiliationsAndNotice{}  % no special notice (required even if empty)


\begin{abstract}
% Reinforcement learning (RL) has achieved notable success in sequential decision-making across high-dimensional perceptual domains such as Atari games, robotic manipulation, and complex simulated environments. However, learning effective policies remains challenging due to low sample efficiency, vulnerability to noise, and poor generalization under partial observability. Existing solutions primarily improve optimization at the training stage, while neglecting the role of network architecture in shaping representation learning and decision-making dynamics. Biological intelligence demonstrate that structural principles embedded in neural circuits can enable efficient, robust learning from sparse, noisy inputs. In particular, the cerebellum offers a compelling model: a shallow feedforward architecture with large-scale feature expansion, sparse connectivity and activation, dual-pathway processing, and dendritic-level modulation for state-dependent computation. We propose a cerebellum-inspired RL architecture that integrate these mechanisms as network-level priors. Large-scale expansion improves representational separability, sparse activation enhances robustness, dual-pathway processing balances fast decision-making with adaptive control, and dendritic modulation enables conditional computation. Experiments on noisy, high-dimensional RL benchmarks show that our approach significantly improves adaptability, robustness, and generalization compared to conventional architectures, highlighting the potential of cerebellar structural priors in advancing reinforcement learning.
Reinforcement learning (RL) has achieved notable performance in high-dimensional sequential decision-making tasks, yet remains limited by low sample efficiency, sensitivity to noise, and weak generalization under partial observability. Most existing approaches address these issues primarily through optimization strategies, while the role of architectural priors in shaping representation learning and decision dynamics is less explored. Inspired by structural principles of the cerebellum, we propose a biologically grounded RL architecture that incorporate large expansion, sparse connectivity, sparse activation, and dendritic-level modulation. Experiments on noisy, high-dimensional RL benchmarks show that both the cerebellar architecture and dendritic modulation consistently improve sample efficiency, robustness, and generalization compared to conventional designs. Sensitivity analysis of architectural parameters suggests that cerebellum-inspired structures can offer optimized performance for RL with constrained model parameters. Overall, our work underscores the value of cerebellar structural priors as effective inductive biases for RL.
\end{abstract}
\section{Introduction}
% Reinforcement learning (RL) has achieved impressive success in sequential decision-making and control, particularly in high-dimensional perceptual domains such as Atari games, robotic manipulation, and complex simulated environments~\cite{anwar2022training,graesser2022state}. However, learning effective policies directly from high-dimensional observations remains a fundamental challenge. RL agents often suffer from low sample efficiency, vulnerability to observation noise, and poor generalization across task variations. These limitations are especially pronounced when policies must be learned from raw sensory inputs under partial observability and stochastic perturbations. 
Reinforcement learning (RL) has shown remarkable progress in sequential decision-making and control across high-dimensional domains such as Atari, robotics, and complex simulations~\cite{anwar2022training,graesser2022state}. Yet, learning effective policies from raw observations remains challenging due to low sample efficiency, sensitivity to noise, and poor generalization issues exacerbated under partial observability and stochastic perturbations.

% A substantial body of prior work has sought to address these challenges through increasingly sophisticated training strategies, including auxiliary objectives, data augmentation, regularization schemes, and large-capacity neural architectures~\cite{shakya2023reinforcement,zhu2023transfer,zeng2024survey}.While effective in certain scenarios, these methods often introduce considerable computational overhead and require extensive hyperparameter tuning. More critically, they primarily improve optimization and data efficiency at the training stage, while paying comparatively less attention to the role of network architecture in shaping representation learning and decision-making dynamics. This raises a fundamental question: Can structural principles embedded within the architecture itself provide a direct and robust means of addressing the challenges in high-dimensional RL?
Existing solutions often rely on advanced training strategies, including auxiliary objectives, data augmentation, regularization, and large-capacity networks~\cite{shakya2023reinforcement,zhu2023transfer,zeng2024survey}. While these can improve optimization and data efficiency, they typically incur high computational cost, require extensive tuning, and pay less attention to how architectural principles shape representation learning and decision-making. This raises a key question: Can network structure itself provide a direct and robust means of addressing the core challenges of high-dimensional RL?

% Biological intelligence offers compelling evidence that the answer may be affirmative. Animals and humans acquire complex behaviors from sparse, noisy, and partially observable sensory inputs, often with minimal explicit reward signals, by leveraging neural circuit structures such as specialized sensory pathways, recurrent dynamics, and adaptive synaptic mechanisms. These systems integrate perception, memory, and action in a temporally coherent manner, enabling rapid generalization across tasks without exhaustive retraining.  
Biological intelligence suggests the answer may be yes: animals and humans learn complex behaviors from sparse, noisy, and partially observable inputs, often with minimal rewards, by leveraging specialized sensory pathways, recurrent dynamics, and adaptive synapses. These neural circuits integrate perception, memory, and action, enabling rapid generalization across tasks without exhaustive retraining.

Recent brain-inspired RL approaches have drawn on models of different brain regions to enhance representation learning, planning, and control. Examples include hippocampus-inspired RL for spatial mapping and episodic memory~\cite{banino2018vector,ritter2018been}, prefrontal cortex inspired RL for hierarchical planning and working memory~\cite{wang2018prefrontal}, and basal ganglia-inspired RL for reward-driven action selection and policy optimization~\cite{joel2002actor,gurney2015new}.

% Among brain regions, the cerebellum stands out for its role in sensorimotor control, predictive modeling, and error correction under high-dimensional, noisy, and temporally delayed inputs~\cite{wolpert1998internal,ito2006cerebellar,zang2023recent}. Recent evidence indicates that climbing fibers (CF) and granule cells (GrC) carry reward-related signals, supporting the view that the cerebellum engages in RL~\cite{wagner2017cerebellar,kostadinov2019predictive}. Its architecture, characterized by precise timing, adaptive learning rules, and efficient feedforward processing, makes it a promising inspiration for RL agents that require rapid adaptation and stable performance in dynamic environments.
The cerebellum plays a key role in sensorimotor control, predictive modeling, and error correction under high‑dimensional, noisy, and delayed inputs~\cite{wolpert1998internal,ito2006cerebellar,zang2023recent}. Recent findings show that climbing fibers (CFs) and granule cells (GrCs) convey reward signals, suggesting cerebellar involvement in RL~\cite{wagner2017cerebellar,kostadinov2019predictive}. Its precise timing, adaptive learning rules, and efficient feedforward processing make it a strong model for RL agents requiring rapid adaptation and stable performance in dynamic environments.

\begin{figure*}[!t]  % t = page top
\centering
\includegraphics[width=0.98\textwidth]{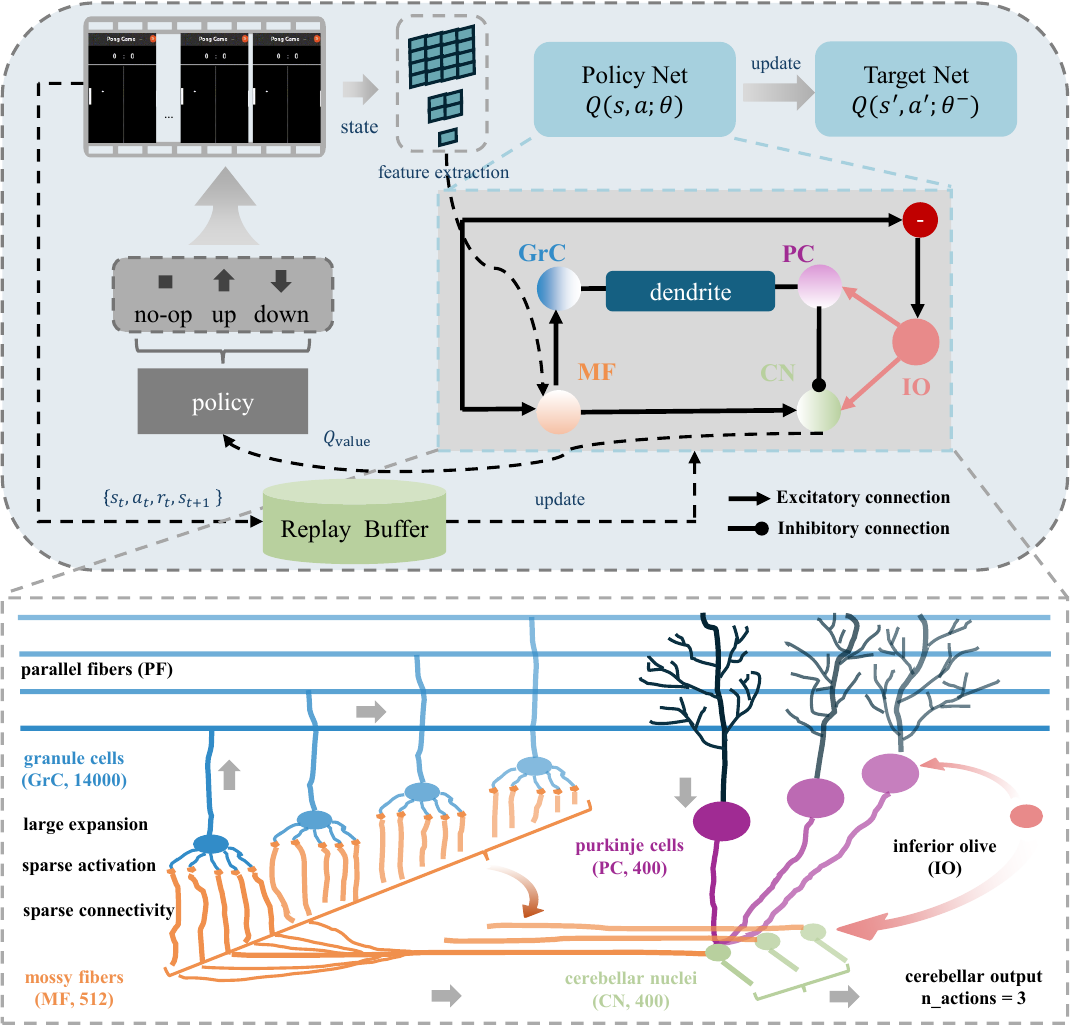} 
\caption{Architecture of the RL framework inspired by
cerebellar circuits and dendritic computational strategies.}
\label{fig1}
\end{figure*}

The main information path in the cerebellum exhibits a stereotyped feedforward architecture (Fig.~\ref{fig1}), mossy fiber (MF) $\rightarrow$ GrC $\rightarrow$ purkinje cell (PC) $\rightarrow$ cerebellar nuclei (CN)), large expansion ($n_{\mathrm{GrC}}:n_{\mathrm{MF}} \approx 27:1$), sparse connectivity (each GrC connects to five MFs), sparse activation (top-k selection) in GrCs, and dendritic-level modulation in PCs that enables state-dependent gating and context-sensitive learning.
From a computational perspective, these principles may directly address RL challenges:
\begin{itemize}
    % \item  \textbf{Dual-pathway architectures} balance low-latency decision-making with high-capacity adaptive control.
    \item \textbf{Large expansion} improves representational separability, aiding downstream value estimation and policy learning.
    \item  \textbf{Sparse activation and connectivity} reduce overlap between patterns and suppress irrelevant or noisy features, enhancing robustness in high-dimensional spaces.
    \item  \textbf{Dendritic modulation} enables state-dependent gating, supporting conditional computation and adaptability under noise and task variation.
\end{itemize}
Together, these mechanisms form a powerful set of architectural inductive biases for RL, offering an alternative to conventional deep network designs. In this work, we propose an RL framework inspired by cerebellar circuits and dendritic computational strategies (CDRL). The architecture incorporates large expansion, sparse connectivity, sparse activation, and dendritic-level modulation as network-level priors. We instantiate this architecture within the Double Deep Q-Network (DDQN) framework by replacing the standard Q-network with a cerebellum-inspired value function approximator. We systematically evaluate the architectural features effects on the model’s robustness and generalization in noisy, high-dimensional RL environments, and demonstrate significant improvements over existing architectures.

\section{Related Work}
\textbf{Algorithmic advances in RL} 

A substantial body of research has sought to improve RL performance through algorithmic optimization and training-centric strategies. One widely explored approach focuses on optimization, aiming to refine optimization dynamics and enhance representation quality. Data augmentation has been extensively applied to improve policy robustness, with examples including automatic augmentation selection~\cite{raileanu2021automatic} and feature-level frame stacking with shifted overlays~\cite{song2024simple}. Auxiliary objectives have been introduced to enrich state and action representations, such as temporal contrastive learning~\cite{zheng2023texttt} and dual-agent reward shaping~\cite{ma2024reward}. Regularization techniques have also been employed to stabilize training, such as Parseval regularization for weight orthogonality~\cite{chung2024parseval} and policy-constrained updates in offline settings~\cite{ran2023policy}. 

Another line of research investigates training methodologies that expand model capacity and enable efficient large-scale learning. Scaling model capacity has proven effective in enhancing representation quality and task performance~\cite{bai2024drpc}, while distributed training facilitates efficient large-scale policy optimization~\cite{ma2024efficient}. 

% Despite these advances, most approaches remain centered on training dynamics, often introducing additional computational overhead and heightened sensitivity to hyperparameter selection.
Despite these advances, most approaches remain centered on training dynamics, often introducing computational overhead and heightened sensitivity to hyperparameter selection.

\textbf{AI Architectures Inspired by Biological Circuits} 

% Biologically inspired neural architectures have profoundly influenced and been widely applied in domains such as image classification, motor learning, and robotic arm manipulation.
Biologically inspired neural architectures have profoundly influenced and been widely applied in fields including image classification, motor learning, and robotic arm manipulation.

Visual Processing Pathways: The hierarchical processing observed in the primate ventral visual stream (the “what” pathway), first characterized by Hubel and Wiesel, has directly influenced the design of convolutional neural networks (CNNs)~\cite{hubel1959receptive}. Subsequent experiments have demonstrated strong similarities between CNN architectures and the ventral visual stream, with state-of-the-art CNN models achieving near-human-level object recognition performance~\cite{yamins2016using}.

Motor Control and Learning: Inspired by the cerebellum’s role in motor coordination, cerebellar trajectory modules are integrated into robotic manipulation systems. These models enable precise trajectory execution and adaptive motor performance, leading to improved control in dynamic environments~\cite{abadia2021cerebellar,zhang2022cerebellum}.

Dendritic Neuron Models: Dendrites have the potential of implementing more complicated computations~\cite{zang2021cellular}. Dendritic branches allow task-dependent gating of relevant features~\cite{veness2021gated}, while branched neuron designs facilitate rapid adaptation in high-dimensional input spaces~\cite{acharya2022dendritic}. More recent studies incorporate nonlinear dendritic gating mechanisms to dynamically integrate information, improving learning efficiency and training stability~\cite{zhang2024lightweight,chavlis2025dendrites}.

\textbf{Brain-Inspired RL} 

Existing brain-inspired studies mainly enhance RL by abstracting neural systems into specific cognitive or functional roles. Representative examples include hippocampus-inspired models for spatial representation and episodic memory~\cite{banino2018vector,ritter2018been}, prefrontal cortex-inspired architectures for hierarchical planning and working memory~\cite{wang2018prefrontal}, and basal ganglia-inspired models for reward-driven action selection and policy optimization~\cite{joel2002actor,gurney2015new}. These models highlight the utility of functional abstractions but place relatively less emphasis on architectural priors.

% In contrast, existing cerebellar-inspired models have typically been developed for relatively simple RL tasks and often employ simplified representations of cerebellar circuitry. 

% In contrast, existing cerebellar-inspired models typically interpret climbing fiber (CF) signals as reward prediction errors to drive synaptic plasticity for error correction and adaptation~\cite{kostadinov2022reward,li2022td,jin2025reward,hoang2025predictive}, while often relying on simplified cerebellar circuit representations. Despite their value in explaining experimental findings under specific learning protocols, the systematic incorporation of detailed cerebellar structural principles as architectural inductive biases into RL remains largely unexplored. Functionally, the cerebellum satisfies core RL requirements, including state representation, action selection, and evaluative feedback processing~\cite{swain2011cerebellum,kuriyama2025theory}.

Previous cerebellum-inspired RL models have attempted to incorporate cerebellar components as auxiliary modules within conventional RL frameworks~\cite{li2022td, hoang2025predictive}, often relying on simplified cerebellar circuits. Some spiking-based RL models implement biologically grounded learning~\cite{masetty2021cerebellum, kuriyama2025theory}, but are constrained by the complexity of training and the limited use of cerebellar structural principles, restricting their scalability and performance in complex tasks. Although the cerebellum has been shown to support key RL operations, including state representation, action selection, and evaluative feedback processing~\cite{swain2011cerebellum, kuriyama2025theory}, existing cerebellum-inspired approaches rarely leverage its structural organization as architectural priors, thereby limiting their potential for effective representation and decision-making in deep RL.

\section{Preliminary and Methodology}
\subsection{RL Framework}
\paragraph{DDQN Algorithm.}
% In our framework, we adopt DDQN~\cite{van2016deep} which mitigates the overestimation bias inherent in standard DQN~\cite{mnih2015human} by explicitly decoupling \emph{action selection} and \emph{action evaluation} through two separate networks: an online network parameterized by $\theta$ and a target network parameterized by $\theta^{-}$.
% In our framework, we adopt DDQN~\cite{van2016deep}, which mitigates the overestimation bias of standard DQN~\cite{mnih2015human}.  
% DQN methods learn an action-value function \(Q(s,a)\), which estimates the expected cumulative reward for taking action \(a\) in state \(s\) and following a given policy \(\pi\) thereafter. Here, the policy \(\pi\) is the mapping from states to actions, which is implicitly induced by the learned Q-function (e.g., via an \(\epsilon\)-greedy strategy). DDQN decouples action selection and action evaluation using two separate Q-networks: an online network parameterized by \(\theta\) and a target network parameterized by \(\theta^{-}\).
In our framework, we adopt DDQN~\cite{van2016deep}, which mitigates the overestimation bias of standard Deep Q-Network (DQN)~\cite{mnih2015human}.  
DQN methods learn a a parameterized action-value function \(Q(s,a;\theta)\), 
which estimates the expected cumulative reward (\(Q_{value}\)) for taking action \(a\) in state \(s\) and following a given policy \(\pi\) thereafter. Here, the policy \(\pi\) maps states to actions, which is implicitly induced by the learned Q-function. In practice, this Q-function is approximated by a neural network. DDQN decouples action selection and action evaluation using two separate Q-networks: an online network parameterized by \(\theta\) and a target network parameterized by \(\theta^{-}\).

The input state is first processed by a three-layer CNN to extract compact spatiotemporal representations. These representations are then fed into the DDQN value estimation module to compute action-value functions. As an off-policy algorithm, DDQN leverages experience replay to learn from past interactions, which further stabilizes training and improves data efficiency (Fig.~\ref{fig1}).

Given an input state $s$, the online network estimates the action-value function $Q(s,a;\theta)$, which represents the expected return of executing action $a$ under the current policy. A greedy policy $\pi(s)=\arg\max_a Q(s,a;\theta)$ is implicitly derived from these estimates and used for action selection during learning and evaluation.

Training is performed by minimizing the mean-squared temporal-difference (TD) error over transitions sampled from the replay buffer $\mathcal{D}$:
% \begin{equation}
% L(\theta) = \mathbb{E}_{(s,a,r,s') \sim \mathcal{D}}
% \Big[ \big( y - Q(s,a;\theta) \big)^2 \Big],
% \end{equation}
% where the target value $y$ is defined as
% \begin{equation}
% y = r + \gamma \,
% Q\big( s', \arg\max_{a'} Q(s',a';\theta); \theta^{-} \big).
% \end{equation}
\begin{equation}
\mathcal{L}(\theta) =
\mathbb{E}_{(s_t,a_t,r_t,s_{t+1}) \sim \mathcal{D}}
\Big[
\big( y_t - Q(s_t,a_t;\theta) \big)^2
\Big],
\end{equation}
where the target value $y_t$ is defined as
\begin{equation}
y_t =
r_t + \gamma \,
Q\big(
s_{t+1},
\arg\max_{a'} Q(s_{t+1},a';\theta);
\theta^{-}
\big).
\end{equation}
The next action is selected by the online network, while its corresponding value is evaluated using the target network, leading to more stable value estimation.
% \paragraph{DDQN Algorithm.}
% We adopt the DDQN algorithm~\cite{van2016deep} as the underlying reinforcement learning framework. DDQN addresses the overestimation bias of standard DQN~\cite{mnih2015human} by decoupling action selection and action evaluation through two networks: an online network with parameters $\theta$ and a target network with parameters $\theta^{-}$.

% The training minimizes the mean squared temporal-difference (TD) error over transitions sampled from the replay buffer $\mathcal{D}$:
% \begin{equation}
% L(\theta) = \mathbb{E}_{(s,a,r,s') \sim \mathcal{D}} 
% \Big[ \big( y - Q(s,a; \theta) \big)^2 \Big],
% \end{equation}
% where the target value $y$ is defined as
% \begin{equation}
% y = r + \gamma \, Q\big( s', \arg\max_{a'} Q(s', a'; \theta); \theta^- \big).
% \end{equation}
% % Here, the next action is selected using the online network, while its value is evaluated using the target network, leading to more stable learning.
% The input states are first processed by a three-layer convolutional neural network (CNN), based on the original DQN architecture with some modifications in layer parameters, to extract spatial and temporal features. These features are then used by the DDQN framework, where the next action is selected using the online network and its value evaluated by the target network. As an off-policy method, DDQN enables learning from past experiences stored in the replay buffer (Fig.~\ref{fig1}), which reduces overestimation bias and improves learning stability.

\paragraph{Q-Network.}
We enhance the standard Q-network in DDQN with a cerebellum-inspired architecture for action-value estimation. This design improves sample efficiency and robustness of value estimation in high-dimensional perceptual control tasks. A dendritic-level modulation mechanism adopted to  allow state-dependent regulation, adaptively controlling information flow under observational noise or task variations. All modifications remain fully compatible with the DDQN optimization objective.
% The network produces action-value estimates for all actions in a given state, from which a greedy policy is implicitly derived for action selection. The architecture introduces four biologically motivated components: large-scale feature expansion, sparse activation, sparse connectivity and a dual-pathway decision structure pathway. 

% \paragraph{Q-Network.}
% Building upon the DDQN framework, we replace the standard Q-network with a cerebellum-inspired architecture. Specifically, the function $Q(s,a;\theta)$ is parameterized by a network that integrate large-scale feature expansion, sparse activation, sparse connectivity, and a dual-pathway decision structure. These components introduce biologically motivated structural inductive biases, enabling more efficient and robust value estimation in high-dimensional perceptual control tasks.

% Furthermore, we incorporate a dendritic-level modulation mechanism to enable state-dependent regulation of high-dimensional sparse pathways. This mechanism allows the Q-network to adaptively modulate information flow under observational noise and task variations, while remaining fully compatible with the DDQN optimization objective described above.

\subsection{The cerebellar architecure in the framework}
% The proposed Cerebellum-inspired Reinforcement Learning (CIRL) model architecture is illustrated in Figure~\ref{fig1}. 
% The model takes a high-dimensional perceptual state \(s_t\) (e.g., stacked grayscale frames) as input 
% and outputs a policy distribution \(\pi(a_t \mid s_t)\) over actions \(a_t\) via the cerebellum-inspired module, 
% which performs feature expansion and conditional modulation. 
% The overall architecture consists of four key components: 
% large-scale feature expansion (granule cell layer), sparse activation and sparse connectivity, 
% dual-pathway decision structure, and dendritic-level modulatory mechanism, 
% which embed structural inductive biases into the reinforcement learning framework. 
% This architecture can be trained in conjunction with standard policy networks such as DQN or DDQN 
% to optimize the action-value function \(Q(s_t, a_t)\) or the policy \(\pi(a_t \mid s_t)\).

% The overall architecture of the proposed CDRL model is illustrated in Figure~\ref{fig1}. 
% The model takes a high-dimensional perceptual state $s_t$ (e.g., stacked grayscale frames) as input and produces a policy distribution $\pi(a_t \mid s_t)$ over actions through a cerebellum-inspired module that performs \emph{large-scale feature expansion} and \emph{state-dependent modulation}.

% In our framework, a high-dimensional perceptual state $s_t$ input (e.g., stacked grayscale frames) is fed into the cerebellar circuit, which outputs a policy distribution $\pi(a_t \mid s_t)$ through large expansion and state-dependent modulation.
In our framework, a high-dimensional perceptual state $s_t$ input is fed into the cerebellar circuit, which outputs a policy distribution $\pi(a_t \mid s_t)$ through large expansion and state-dependent modulation.

% The cerebellar architecture consists of three core components:
% (i) large expansion implemented via GrC layer,
% (ii) sparse MF-GrC connectivity to induce pattern separation, 
% (iii) sparse activation in GrCs to promot efficient coding of input features,
% (iv) and a dual-pathway decision structure for balancing capacity and responsiveness.
% and (iv) a dendritic-level modulatory mechanism that enables conditional regulation of high-dimensional sparse pathways.
The cerebellar architecture consists of three core components: (i) large expansion implemented via GrC layer, (ii) sparse MF-GrC connectivity, and (iii) sparse activation in GrCs. 
Together, these components embed biologically inspired structural inductive biases into the RL framework, the computational procedure of the cerebellar architecture is summarized in Algorithm~\ref{alg:cirl} of the Appendix.

The proposed architecture is designed as a modular and algorithm-agnostic component, and can be trained jointly with standard RL algorithms such as DQN or DDQN to optimize either the action-value function, from which the policy $\pi$ is implicitly derived.

In our framework, the cerebellar GrCs map the input state from MFs to a high-dimensional feature space. 
Specifically, given an input vector \(x_t \in \mathbb{R}^{d_s}\), where \(d_s\) denotes the dimensionality of the input, a sparse random projection \(\Phi: \mathbb{R}^{d_s} \to \mathbb{R}^{d_{GrC}}\) generates sparse activations:
% Specifically, given an input vector \(x_t\), a sparse random projection 
% \(\Phi: \mathbb{R}^{d_s} \to \mathbb{R}^{d_{gc}}\) generate sparse activations:
\begin{equation}
\mathbf{h}_{GrC} = \text{ReLU}(\Phi(x_t)) \odot \mathbf{v}, \quad 
\mathbf{v} \sim \text{Bernoulli}(p),
\end{equation}
where \(d_{GrC} \gg d_s\) and \(\mathbf{v}\) is a sparsity mask ensuring most neurons are inactive, 
enhancing representational capacity and improving generalization.
% \subsubsection{Sparse Mechanisms}
In the cerebellar circuit, each GrC receives inputs from only a small subset of MFs, enforcing structural sparsity to promote pattern separation. Additionally, the GrCs are sparsely activated to facilitate the pattern separation and information capacity. Subsequently, GrCs project to PCs, which target the cerebellar output cells in CN. Because of sparsely activated GrCs,, each PC integrates inputs from only a subset of GrC features, forming a local sparse projection:
\begin{equation}
\mathbf{h}_{PC} = \mathbf{W}_{GrC\_PC} \, \mathbf{h}_{GrC},
\end{equation}
where the synaptic weight matrix $\mathbf{W}_{GrC\_PC}$ is sparse.
Together, sparse connectivity and sparse activation reduce information redundancy, suppress noise propagation, and preserve salient features, providing stable and discriminative inputs for downstream decision-making units.
% \subsubsection{Dual-Pathway Decision Structure}
In our model, the final output of the CN is computed by the weighted combination of the direct and indirect pathways:
\begin{align}
\mathbf{CN}_{direct} &= f_{MF\_CN}(x_t), \\
\mathbf{CN}_{indirect} &= f_{PC\_CN}(\mathbf{h}_{PC}).
\end{align}
% Subsequently, the cerebellar output is fed into the Q-network to perform reinforcement learning. Training follows the standard DDQN framework:
Subsequently, the cerebellar output is fed into the Q-network for value estimation. The overall training objective follows the standard DDQN formulation defined in Eq.~(1). Specifically, the Q-function is parameterized by the proposed cerebellum-inspired network, and optimized by minimizing $\mathcal{L}(\theta)$. $\theta$ denotes the set of trainable parameters, including the GrC$\rightarrow$PC synaptic weights and the fusion weights.
The introduction of the cerebellar module does not alter the underlying optimization procedure. Training follows the standard DDQN paradigm with experience replay and $\epsilon$-greedy exploration.

% and the final CN output is a weighted combination of the two pathways.
% The dual-pathway design accelerate learning convergence and enhances adaptability to diverse task environments.
% After processing by the aforementioned modules, the CN output is fed into a policy or Q-network to perform reinforcement learning. 
% Training follows the standard DDQN framework:
% \begin{multline}
% \mathcal{L}(\theta) = \mathbb{E}_{(s_t, a_t, r_t, s_{t+1})} 
% \Big[ \big( r_t + \gamma \max_{a'} Q_{\theta^-}(s_{t+1}, a') \\
% - Q_\theta(s_t, a_t) \big)^2 \Big].
% \end{multline}
% where $\theta$ denotes the set of trainable parameters, including the GrC$\rightarrow$PC synaptic weights and the fusion weights. Standard training techniques, such as experience replay and $\epsilon$-greedy exploration, are employed, consistent with conventional DQN training.

% where \theta denotes the set of trainable parameters, including the GrC→PC weights and fusion weights. Standard training techniques such as experience replay and \(\epsilon\)-greedy exploration are used, consistent with conventional DQN training.
% Standard training techniques such as experience replay and \(\epsilon\)-greedy exploration are used, consistent with conventional DQN training.

\subsection{Dendritic Modulation Mechanism}
The computational procedure of the dendritic mechanism, mimicking gating function in PC complex dendrites, is summarized in Algorithm~\ref{alg:dendritic-gate} of the Appendix. We implement dendritic mechanisms as a population-driven, non-trainable gain module on the GrC–PC pathway, abstracting sub-branch integration via random hyperplane projections and top-k selection, with slow exponential moving average (EMA) dynamics regulating excitability. This modulates signal amplitude without altering synaptic weights or requiring backpropagation.

% \begin{algorithm}[tb]
% \caption{Dendritic Modulation Mechanism}
% \label{alg:dendritic-gate}
% \begin{algorithmic}
%    \STATE {\bfseries Input:} GrC activations $h_{gc}$, dendritic parameters $\Theta_d$
%    \STATE {\bfseries Output:} modulated activations $x_{\text{mod}}$, global\_gain
%    \STATE $s \gets \text{Aggregate}(h_{gc})$
%    \STATE $b \gets \text{SelectBranches}(s;\, \Theta_d)$
%    \STATE $g_{\text{inst}} \gets \text{DendriticGate}(b)$
%    \STATE $g \gets \text{TemporalIntegrate}(g_{\text{inst}})$
%    \STATE $x_{\text{mod}} \gets x \odot \text{Gain}(g)$
%    \STATE $global\_gain \gets \text{GlobalGain}(g)$
%    \STATE \textbf{return} $x_{\text{mod}}, global\_gain$
% \end{algorithmic}
% \end{algorithm}

% Given sparse GrC activations $\mathbf{G}_t \in \mathbb{R}^{B \times D}$ at time $t$,
Given sparse GrC activations $\mathbf{G}_t \in \mathbb{R}^{B \times D}$ at time $t$, where $B$ denotes the batch size and $D$ denotes the number of GrC, and $b \in \{1,\dots,B\}$ indexes individual samples within the batch, we compute a population-level representation as:
\begin{equation}
\bar{\mathbf{g}}_t
=
\frac{1}{B}
\sum_{b=1}^{B}
\mathbf{G}_t^{(b)},
\quad
\mathbf{G}_t \in \mathbb{R}^{B \times D}.
\end{equation}

The population activity is $\ell_2$-normalized as:
\begin{equation}
\tilde{\mathbf{g}}_t
=
\frac{\bar{\mathbf{g}}_t}{\|\bar{\mathbf{g}}_t\|_2}.
\end{equation}

% Each dendritic branch is associated with a fixed random hyperplane~\cite{veness2021gated},
% parameterized by a unit normal vector $\mathbf{h}_m \in \mathbb{R}^{D}$ with
% $\|\mathbf{h}_m\|_2 = 1$, which is introduced to project the GrC outputs without
% participating in backpropagation, thereby reducing the training cost.

Each dendritic branch $m$ employs a fixed random hyperplane~\cite{veness2021gated}
with unit normal $\mathbf{h}_m \in \mathbb{R}^{D}$ ($\|\mathbf{h}_m\|_2=1$),
which projects GrC outputs via
\begin{equation}
p_{t,m} = \tilde{\mathbf{g}}_t^\top \mathbf{h}_m,
\end{equation}
where $M$ denotes the number of dendritic branches.
The hyperplanes are fixed and excluded from backpropagation, thereby reducing the training cost.

% Each dendritic branch is modeled as a fixed random hyperplanes~\cite{veness2021gated}, which are introduced to project the GrC outputs without participating in backpropagation, thereby reducing training cost:
% $\mathbf{h}_m \in \mathbb{R}^{D}$ with $\|\mathbf{h}_m\|_2=1$.

% The dendritic projections are given by:
% \begin{equation}
% p_{t,m}
% =
% \tilde{\mathbf{g}}_t^\top \mathbf{h}_m,
% \quad m=1,\dots,M,
% \end{equation}
% where $M$ denotes the number of dendritic branches.

Only the top-k dendritic branches are activated,
\begin{equation}
\mathcal{S}_t
=
\text{TopK}\left(\{p_{t,m}\}_{m=1}^{M}, K\right),
\quad K=\rho M.
\end{equation}

The dendritic integration is defined as:
\begin{equation}
\mathbf{d}_t
=
\sum_{m \in \mathcal{S}_t}
\sigma(\beta p_{t,m})\,\mathbf{h}_m,
\end{equation}

yielding the instantaneous dendritic signal:
\begin{equation}
\mathbf{z}_t = \sigma(\mathbf{d}_t).
\end{equation}

To introduce a slow temporal scale, dendritic signals are accumulated via:
\begin{equation}
\mathbf{e}_t
=
\tau \mathbf{e}_{t-1}
+
(1-\tau)\mathbf{z}_t,
\quad \tau \in (0,1).
\end{equation}

Finally, the accumulated dendritic state modulates GrC inputs to PCs:
\begin{equation}
\hat{\mathbf{G}}_t
=
\mathbf{G}_t
\odot
\left(
\mathbf{1}
+
\alpha(\mathbf{e}_t - 0.5)
\right).
\end{equation}

Equations (7)--(14) define a dendritic modulation mechanism that operate on
population-level GrC statistics, integrate information across multiple random
dendritic branches, and accumulates modulatory signals over a slow temporal
scale, while remaining fully decoupled from gradient-based learning.
\section{Experiments}
\subsection{Experimental Setup}
\begin{figure}[!htb]
\centering
\includegraphics[width=0.15\textwidth, keepaspectratio]{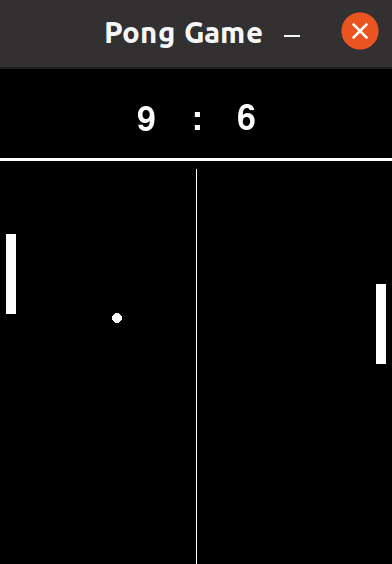}
\caption{Self-built Pong game environment.}
\label{fig-pong}
\end{figure}

Existing implementations lack the flexibility necessary to support interactive environments with configurable physical dynamics. Following common practice in prior works~\cite{adhikari2021rl,anwar2022training,takano2024robustness,chen2025foveal}, to address this limitation, we develop a custom Pong environment (Fig.~\ref{fig-pong}) tailored to our experimental requirements which is inspired by Atari Pong. (key parameters in Table~\ref{tab:pong_env_params_3col} of the Appendix).

% The key parameters of our pong game environment provided in Table~\ref{tab:pong_env_params_3col} of the Appendix.

% The environment incorporate frame stacking, grayscale preprocessing, and a discrete action space 
% (\emph{up}, \emph{down}, and \emph{no-operation (noop)}), making it suitable for reinforcement learning research in high-dimensional perceptual control tasks. The agent on the left is the trained policy, while the opponent on the right is the built-in game controller, whose behavior is governed by tracking the physical position of the ball. Model training is performed using the discrete Double DQN (DDQN) algorithm, 
% with experience replay and an \(\epsilon\)-greedy exploration strategy employed during training. 

% The training hyperparameters, including the learning rate, discount factor \(\gamma\), 
% replay buffer capacity, and batch size, are summarized in the appendix. 
% Evaluation metrics primarily include the average return, win rate, 
% and performance degradation under observation noise or action perturbations, 
% which are used to assess model robustness. 
% Generalization ability is evaluated by measuring performance across environment variations or parameter perturbations. Unless otherwise specified, all results are averaged over multiple 
% runs with fixed 5 seeds to reduce variance.

The environment incorporates frame stacking, grayscale preprocessing, and a discrete action space (\textit{up}, \textit{down}, \textit{no-operation}). The left agent follows our trained policy, while the right agent is a built-in controller that tracks and controls the ball's position. Training is conducted using the discrete DDQN with experience replay and $\epsilon$-greedy exploration. Hyperparameters, including learning rate, discount factor ($\gamma$), replay buffer size, and batch size, are provided in Table~\ref{tab:pong_train_params_3col} of the Appendix. Rewards are assigned +1 for scoring, -1 for conceding, and 0 otherwise, with each game played to 21 points, yielding a sparse episodic signal. Evaluation metrics include win rate, average reward, robustness under observation noise or action perturbations, and generalization performance across environment variations. Unless stated otherwise, results use five fixed seeds.
% Unless otherwise specified, results are averaged over five fixed random seeds.

% To ensure fair comparison, we design a baseline model that shares the same feature-extraction backbone and feedforward network depth as the proposed CDRL model. The key difference is that CDRL incorporates sparse connectivity and sparse activation, to reduce parameter number and training cost while achieving higher accuracy and rewards, but the baseline model does not employ any sparsity mechanisms. 
To ensure fair comparison, we design a baseline model that shares the same feature-extraction backbone and feedforward network depth as the proposed CDRL model. The key difference is that CDRL incorporates sparse connectivity and sparse activation, reducing the parameter count to $\sim 6$ M (compared with $\sim 13$ M for the baseline) and training cost, while achieving higher accuracy and rewards, but the baseline model does not employ any sparsity mechanisms.

\paragraph{Sample Efficiency Evaluation.}
Sample efficiency is evaluated by measuring learning progress against the number of environment interaction steps rather than episodes.
To obtain a stable estimate of performance under the high variance and non-stationary dynamics of RL, we apply EMA to the reward sequence recorded at each training step.
EMA assigns larger weights to more recent observations while retaining information from earlier samples.

% , thereby providing a smooth yet responsive characterization of learning dynamics.

Formally, let $r_t$ denote the episode reward observed at training step $t$.
The EMA-smoothed reward $\hat{r}_t$ is computed recursively as
\begin{equation}
\hat{r}_t = \alpha r_t + (1 - \alpha)\hat{r}_{t-1},
\end{equation}
where $\alpha \in (0, 1]$ controls the decay rate of past observations.
A larger $\alpha$ emphasizes recent rewards, while a smaller value yields stronger smoothing over longer horizons.

\paragraph{Robustness Evaluation.}
To evaluate the robustness of our framework, we introduce controlled perturbations during evaluation while keeping the training environment unchanged. 
Specifically, Gaussian observation noise~\cite{zhang2020robust} is added to the agent’s perceptual input:
\begin{equation}
\tilde{s}_t = s_t + \epsilon_t, \quad \epsilon_t \sim \mathcal{N}(0, \sigma^2),
\end{equation}
where \(s_t\) denotes the original observation and \(\tilde{s}_t\) is the corrupted observation. 
In addition, random action perturbations~\cite{tessler2019action,lyle2022learning} are applied to simulate execution noise under realistic conditions, 
where the selected action is stochastically replaced with a random action with a fixed probability. 

Notably, both observation noise and action perturbations are applied only to the evaluated agent. 
The opponent follows a deterministic position-tracking policy and is not affected by these perturbations.
% making the evaluation setting strictly more challenging. 
Under this asymmetric configuration, we focus on comparing the relative robustness gains 
of the cerebellum-inspired model and the dendritic-augmented model with respect to the baseline. 
This design ensures that observed robustness improvements arise from the intrinsic structural advantages 
of the proposed architectures rather than from reduced task difficulty.

We further evaluate the models under a sticky action noise setting, with a stickiness parameter of $0.25$, which introduces a probability that the agent’s previous action is repeated in the current step. Sticky noise increases action execution uncertainty, simulating the challenge of precise control and providing a rigorous test of policy robustness.
% We further evaluated the models under a sticky action noise setting, with a stickiness parameter of $0.25$, which introduces a probability that the agent's previous action is repeated in the current step. Sticky noise simulates action execution uncertainty, thereby increasing the challenge of precise control and testing the robustness of learned policies.

\paragraph{Generalization Evaluation.}
As shown in Table~\ref{tab:generalization}, generalization performance is assessed by modifying predefined environment parameters at test time, including the ball velocity, decomposed into horizontal and vertical components (Ball Spd. (H) and Ball Spd. (V)), as well as the paddle length (Paddle Len.) and paddle movement speed (Paddle Spd.). These variations induce systematic changes in environment dynamics while preserving the task objective. The agent’s performance on unseen configurations evaluates its generalization beyond the training distribution.
\begin{table}[!htb]
\caption{Environment parameters used for evaluating model generalization.}
\label{tab:generalization}
\centering
\resizebox{1\linewidth}{!}{ 
\setlength{\tabcolsep}{3.5pt}      % 原来 4pt，轻微缩小
\renewcommand{\arraystretch}{1.05} % 原来 1.1，轻微压缩
\fontsize{9}{11}\selectfont        % 保持 9pt，不动
\begin{tabular}{c c c c c}
\toprule
Param. & Ball Spd. (H) & Ball Spd. (V) & Paddle Len. & Paddle Spd. \\
\midrule
Train  & 12 & 8  & 80 & 5 \\
Test 1 & \cellcolor{ball_speed}15 & \cellcolor{ball_speed}10 & 80 & 5 \\
Test 2 & \cellcolor{ball_speed}18 & \cellcolor{ball_speed}12 & 80 & 5 \\
Test 3 & 12 & 8  & \cellcolor{paddle_length}60 & 5 \\
% Test 4 & 12 & 8  & \cellcolor{paddle_length}40 & 5 \\
Test 4 & 12 & 8  & \cellcolor{paddle_length}20 & 5 \\
% Test 6 & 12 & 8  & 80 & \cellcolor{paddle_speed}1 \\
Test 5 & 12 & 8  & 80 & \cellcolor{paddle_speed}2 \\
Test 6 & 12 & 8  & 80 & \cellcolor{paddle_speed}3 \\
Test 7 & 12 & 8  & 80 & \cellcolor{paddle_speed}4 \\
\bottomrule
\end{tabular}
}
\end{table}
% \begin{table}[!htb]
% \caption{Generalization settings.}
% \label{tab:generalization}
% \centering
% \setlength{\tabcolsep}{4pt}        % 缩小列间距（默认 6pt）
% \renewcommand{\arraystretch}{1.1}  % 稍微压缩行距
% \fontsize{9}{11}\selectfont        % 明确 9pt，符合会议习惯
% \begin{tabular}{c c c c c}
% \toprule
% Param. & Ball Spd. (H) & Ball Spd. (V) & Paddle Len. & Paddle Spd. \\
% \midrule
% Train  & 12 & 8  & 80 & 5 \\
% Test 1 & \cellcolor{ball_speed}18 & \cellcolor{ball_speed}12 & 80 & 5 \\
% Test 2 & \cellcolor{ball_speed}24 & \cellcolor{ball_speed}16 & 80 & 5 \\
% Test 3 & 12 & 8  & \cellcolor{paddle_length}60 & 5 \\
% Test 4 & 12 & 8  & \cellcolor{paddle_length}40 & 5 \\
% Test 5 & 12 & 8  & \cellcolor{paddle_length}20 & 5 \\
% Test 6 & 12 & 8  & 80 & \cellcolor{paddle_speed}1 \\
% Test 7 & 12 & 8  & 80 & \cellcolor{paddle_speed}2 \\
% Test 8 & 12 & 8  & 80 & \cellcolor{paddle_speed}3 \\
% Test 9 & 12 & 8  & 80 & \cellcolor{paddle_speed}4 \\
% \bottomrule
% \end{tabular}
% \end{table}
\subsection{CDRL Model Evaluation}
We first assess sample efficiency to measure the CDRL’s fundamental learning performance, then examine robustness under stochastic perturbations to evaluate stability, and finally evaluate generalization to novel conditions to assess the model’s ability to maintain performance in unseen scenarios.
% We first assess CDRL’s sample efficiency to measure its learning performance, then examine robustness to stochastic perturbations to evaluate stability, and finally evaluate generalization to novel conditions to assess its ability to maintain performance in unseen scenarios.
\subsubsection{Sample Efficiency Testing}
\begin{figure}[!htb]
\centering
\includegraphics[width=0.5\textwidth, keepaspectratio]{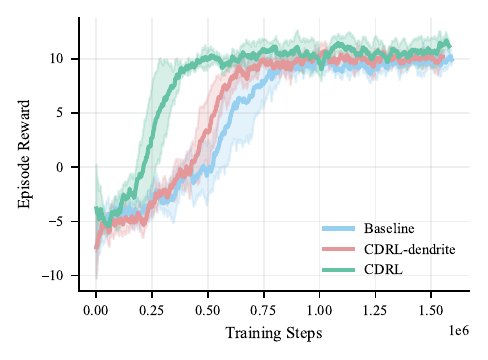}
\caption{Sample efficiency of different models during training, measured by performance versus number of training steps.}
\label{fig-efficient}
\end{figure}

In Figure~\ref{fig-efficient}, we compare the reward dynamics of CDRL, CDRL-dendrite (with dendritic components removed), and the baseline model during training. CDRL achieves a higher steady-state reward more quickly than the other two models, demonstrating superior sample efficiency. In contrast, the reward performance of CDRL-dendrite saturates faster and at a higher level compared to the baseline model. These results indicate that both cerebellum-inspired and dendritic mechanisms contribute to more effective policy acquisition under limited interaction data.
\subsubsection{Robustness Testing}
\begin{figure}[!htb]
\centering
\includegraphics[width=0.50\textwidth, keepaspectratio]{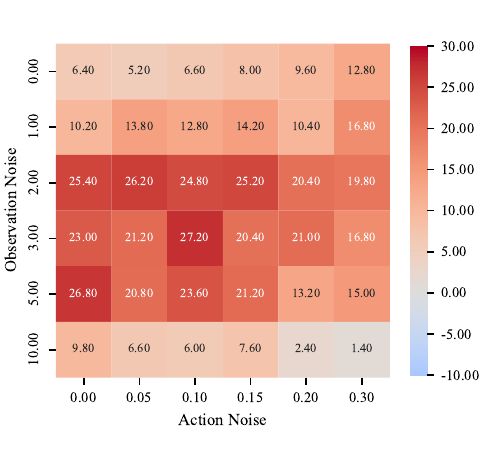}
\caption{Heat map of the difference in robustness between the CDRL model and the baseline model.}
\label{fig-dendrite-baseline-winrate_diff}
\end{figure}

For robustness testing, models are trained in the customized Pong environment and evaluated under stochastic perturbations applied to both observations and actions. Figure~\ref{fig-dendrite-baseline-winrate_diff} illustrates the difference in win rate between the CDRL model and the baseline model. Positive values indicate cases where CDRL outperforms the baseline. Across a wide range of observation noise levels (\(0.00\)–\(10.00\)) and action perturbation probabilities (\(0.00\)–\(0.30\)), CDRL consistently demonstrate superior robustness compared to baseline. The win rates of CDRL and baseline under different combinations of noise parameters are reported in Table~\ref{tab:winrate_wide} of the Appendix. In addition, the CDRL-dendrite model also exhibits superior robustness under most test conditions relative to baseline, suggesting that both the cerebellar architecture and dendritic mechanisms contribute to the robustness improvements. Detailed results are provided in Table~\ref{tab:winrate_cirl} of the Appendix.
% Figure~\ref{fig-cirl-baseline-win rate_diff} illustrate the difference in win rate between the proposed model and the baseline. 
% Positive values (shown in red) indicate cases where the CIRL model outperforms the baseline. 
% Under a wide range of observation noise levels (\(0\)–\(5\)) and action perturbation probabilities (\(0\)–\(0.3\)), the CIRL model consistently demonstrate superior robustness in the majority of test conditions compared to the baseline.

% \begin{figure}[!htb]
% \centering
% \includegraphics[width=0.5\textwidth, keepaspectratio]{fig-cirl-baseline-win rate_diff.pdf}
% \caption{Heat map of the difference in robustness between the CIRL model and the baseline model.}
% \label{fig-cirl-baseline-win rate_diff}
% \end{figure}

Table~\ref{tab:sticky_noise} summarizes the performance of baseline, CDRL-dendrite, and CDRL models under sticky action noise setting. Across multiple trials, the CDRL model consistently achieves the highest win rate and average reward, outperforming both the baseline and CDRL-dendrite models. These results highlight the effectiveness of cerebellum-inspired architecture and the dendritic-level modulation mechanism in enhancing policy robustness under noises.

\begin{table}[!htp]
\caption{Performance under sticky action noise ($p=0.25$).}
\label{tab:sticky_noise}
\centering
\setlength{\tabcolsep}{5.5pt}
\renewcommand{\arraystretch}{1.1}
\fontsize{9}{11}\selectfont
\begin{tabular}{lccc}
\toprule
Model & Baseline & CDRL-dendrite & CDRL \\
\midrule
win rate (\%) & 91.20$\pm$3.54 & 97.20$\pm$1.47 & \textbf{97.60$\pm$1.85} \\
reward & 8.38$\pm$0.85 & 9.71$\pm$0.50 & \textbf{10.07$\pm$1.97} \\
\bottomrule
\end{tabular}
\end{table}

\subsubsection{Generalization Testing}
\begin{figure}[!htb]
\centering
\includegraphics[width=0.5\textwidth, keepaspectratio]{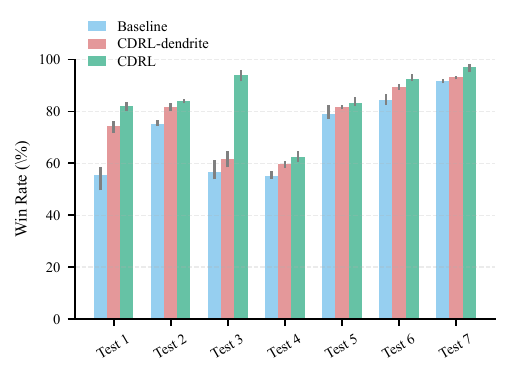}
\caption{Generalization performance of different models under unseen environment configurations.}
\label{fig-generalize}
\end{figure}

As shown in Figure~\ref{fig-generalize}, across three configuration variations: ball speed (tests 1–2), paddle length (tests 3–4), and paddle speed (tests 5–7), CDRL consistently achieves the highest win rate in the test stage, followed by CDRL-dendrite, with the baseline model performing the lowest. These results indicate that both the cerebellar architecture and dendritic modulation contribute to the superior generalization observed in the CDRL model.

\subsection{Sensitivity Analysis of the Cerebellar Architectural Features}
% To systematically evaluate the contributions of different components in CDRL, we conduct a series of ablation experiments. Each experiment selectively removes or modifies a key structural element, such as the GrC layer expansion, sparse machenisms and the dual-pathway decision structure. This approach allows us to quantify the impact of each mechanism on model performance, including sample efficiency, robustness to observation and action noise, and overall learning stability. The results provide insight into which structural priors are most critical for enhancing reinforcement learning in high-dimensional perceptual tasks.

The architecture of CDRL possesses several critical features, including the GrC layer expansion ratio, sparse connection, and sparse activation. In biological circuits, these features may be constrained by developmental cost and functional requirements. However, the parameter settings underlying these features may not be optimal for RL tasks, particularly in high‑dimensional perceptual domains. To quantify the impact of each feature on model performance and guide the design of cerebellum‑inspired architectures for RL, we conduct a series of sensitivity tests by selectively removing or modifying one key feature at a time.

% The cerebellar architecture of our CDRL model posess several critical features, including the GrC layer expansion ratio, sparse connections, and sparse activation. To quantify the impact of each feature on model performance—including sample efficiency, robustness to observation and action noise, and overall learning stability. We conduct a series of sensitivity tests by selectively removing or modifying one key feature at a time. The simulation results provide insights into which structural priors are most critical for enhancing RL in high-dimensional perceptual tasks.

\paragraph{Expansion Ratio.}
% Figure~\ref{fig-auc-expand} and the Figure~\ref{fig-expand1} of the Appendix present the sensitivity test results for varying the number of GrCs in the network. As shown in Figure~\ref{fig-auc-expand}, a larger expansion ratio generally improves performance under moderate levels of observation and action noise. However, this comes with increased computational cost. As shown in the Figure~\ref{fig-expand1} of the Appendix, both the win rate and the average reward increase with the expansion ratio. suggesting that larger expansion ratios enhance feature representation and pattern separation capabilities\cite{zou2025structural}, thereby reliably improving RL performance. However, the benefit of expansion gradually diminishes when the GrC count exceeds 14,000 units. 

% Figure~\ref{fig-auc-expand} presents the sensitivity analysis of the model's AUC for varying the number of GrCs in the network. All tested configurations achieve higher AUC than the baseline. As shown in Figure~\ref{fig-expand1} of the Appendix, both the win rate and the average reward increase with the expansion ratio, suggesting that larger expansion ratios enhance feature representation and pattern separation capabilities~\cite{zou2025structural}, thereby enhancing RL performance. However, the performance gain gradually diminishes when the GrC count exceeds 14,000 units, and larger expansion ratios incur higher computational cost.

\begin{figure}[!htb]
\centering
\includegraphics[width=0.5\textwidth, keepaspectratio]{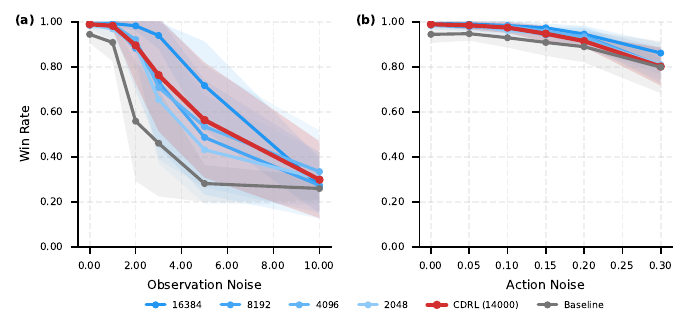}
\caption{Sensitivity results of the expansion ratio. (a) Observation noise. (b) Action noise.
}
\label{fig-auc-expand}
\end{figure}

In Figure~\ref{fig-auc-expand}, all tested model variants achieve a higher win rate than the baseline. The model with 16,384 GrCs demonstrates stronger robustness than CDRL in the medium noise range but shows no advantage in the low or high noise ranges. Given the larger computational load (parameters) associated with the more expanded architecture, the cerebellar design may offer a reasonable trade-off between performance and computational cost. By enhancing feature representation and pattern separation~\cite{zou2025structural}, a modestly large expansion can improve RL performance while maintaining acceptable computational overhead.
% Figure~\ref{fig-auc-expand} presents the sensitivity test results for varying GrC counts, where all tested configurations of the model variants achieve higher win rate compared to the baseline. The model with 16,384 GrCs showing a slight improvement over CDRL (14000 GrCs), indicating that larger expansion ratios can further enhance robustness. As shown in Figure~\ref{fig-expand1} of the Appendix, in the absence of noise, both the win rate and the average reward generally increase with the number of GrCs. However, the performance gain gradually diminishes when the GrC count exceeds 14,000 units. This suggests that the current configuration provides a reasonable trade-off between performance and computational cost, while still benefiting from larger expansion ratios, which enhance feature representation and pattern separation~\cite{zou2025structural}, thereby improving RL performance.
% Figure~\ref{fig-expand1} shows the sensitivity test results regarding the number of Granule Cells (GrC) in the network. All evaluated configurations achieve higher performance than the baseline (dashed lines). And the results suggest that enhanced feature representation and pattern separation capabilities, because of larger expansion ratios\cite{zou2025structural}, in turn improve RL performance. This is demonstrated by the increase in both average reward and win rate(Fig.~\ref{fig-expand1}). Both metrics improve as the number of GrC increases, saturating when the GrC count exceeds 14,000 units. 

% \vspace{-5pt}

\begin{figure}[!htb]
\centering
\includegraphics[width=0.5\textwidth, keepaspectratio]{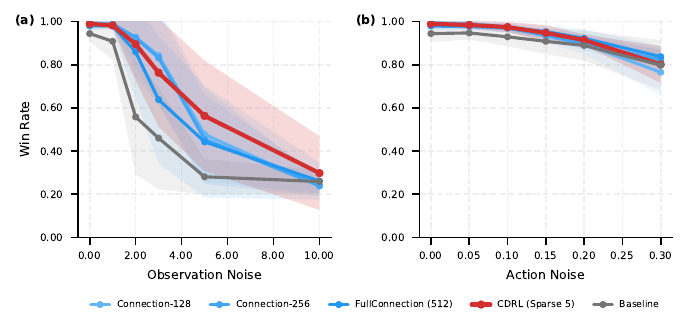}
% \caption{AUC curves of different models based on observation noise.}
\caption{Sensitivity results of the sparse connectivity. (a) Observation noise. (b) Action noise.}
\label{fig-connection}
\end{figure}

\paragraph{Sparse mechanisms.}
In the cerebellar architecture, each GrC receives inputs from only a small subset of MFs. Such sparse projections reduce overlap between different contexts and attenuate observation noise, limiting its propagation and preventing small perturbations from influencing all GrCs. The benefit of sparse connection is minimal in the low‑noise range (Fig.~\ref{fig-connection}a). However, the model with sparse connection demonstrates superior robustness when noise levels are high. Among the evaluated variants, the CDRL model consistently outperforms the baseline (fully connected) across all observation noise levels. As expected, the degree of connection sparsity does not significantly affect model performance under action‑noise conditions (Fig.~\ref{fig-connection}b). Overall, these results suggest that sparse connections should be prioritized due to their lower computational cost (parameters) and superior performance.

% Second, sparse activation at the GC–PC stage is implemented using a winner-take-all (WTA) mechanism, in which only 1\% of GrCs are active at any given time. This selective activation improves the efficiency and discriminability of information encoding, and theoretically mitigates the influence of action noise during learning. As shown in the marginal AUC curves under varying action noise levels (Fig.~\ref{fig-activate}), WTA-based sparse activation improves robustness to stochastic action perturbations. At low action noise levels, the performance of all models decreases only slightly. When the action noise is above 0.15, the CDRL model exhibits a slower rate of performance degradation.
Cerebellar GrCs are sparsely activated, with only a small fraction active at any given time. This property has been shown to be critical in tasks such as continual learning and multitask learning~\cite{zou2025fly,zou2025flylora}. In our sensitivity analysis, the CDRL model consistently outperforms the baseline model (without top‑k). It also lies within a high‑performance range compared to other model variants with denser GrC activation. However, the GrC activation level does not significantly affect model performance under action‑noise conditions. In summary, these results suggest that sparse activation in GrCs should be prioritized due to its overall performance.
% Cerebellar GrCs are sparsely activated, with only a small fraction active at any given time. This property has been demonstrated to be critical in tasks such as continual learning and multitask learning~\cite{zou2025structural}. In our sensitivity analysis under observation noise, all evaluated configurations achieve higher performance than the baseline. However, varying the activation level of GrCs does not consistently improve model performance under either observation noise or action noise, although our model’s win rate remains in the higher range compared to variants with different levels of GrC activation (Fig.~\ref{fig-activate}).
\begin{figure}[!htb]
\centering
\includegraphics[width=0.5\textwidth, keepaspectratio]{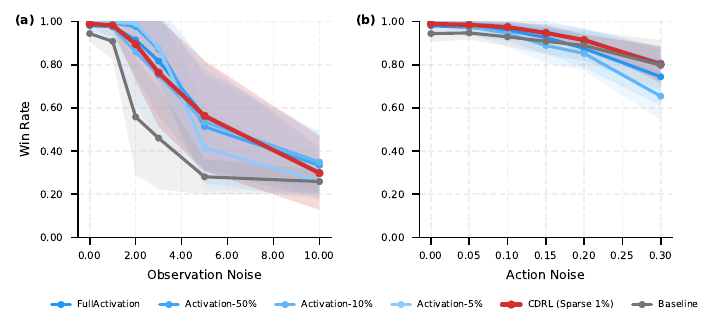}
\caption{Sensitivity results of the sparse activation. (a) Observation noise. (b) Action noise.}
\label{fig-activate}
\end{figure}
\section{Conclusion and Future Work}
In this work, we developed a cerebellum‑based RL model for a high‑dimensional perceptual version of the Pong game with discrete actions. Our results highlight the critical role of the cerebellar architecture and dendritic computation mechanisms in improving win rate, cumulative reward, robustness, and generalization. Sensitivity analysis suggests that the cerebellum provides an optimized structure for RL, although increasing the expansion ratio can enhance performance at the cost of higher computational demand.

Our experiments were limited to the Pong environment with a discrete DDQN framework, constraining task complexity. Continuous‑action tasks and high‑degree‑of‑freedom environments remain unexplored. Future work could investigate the architecture’s transferability, its applicability to continuous control tasks such as robotic manipulation, and its performance in multi‑task learning and knowledge transfer scenarios. Incorporating online plasticity mechanisms or training under high‑noise and delayed‑feedback conditions could further assess and improve robustness and generalization.

Overall, this work demonstrates the potential of cerebellar circuits and dendritic strategies as computational inductive biases in RL and offers promising directions for broader application across more complex tasks.

\section*{Impact Statement}
By integrating cerebellum‑inspired structural priors into reinforcement learning architectures, we achieve significant improvements in sample efficiency, robustness, and generalization on noisy, high‑dimensional tasks with minimal computational overhead.  This work highlights the potential of biologically grounded architectural design as an effective inductive bias for advancing deep RL.

\bibliography{cirl_reference}
\bibliographystyle{icml2026}

%%%%%%%%%%%%%%%%%%%%%%%%%%%%%%%%%%%%%%%%%%%%%%%%%%%%%%%%%%%%%%%%%%%%%%%%%%%%%%%
%%%%%%%%%%%%%%%%%%%%%%%%%%%%%%%%%%%%%%%%%%%%%%%%%%%%%%%%%%%%%%%%%%%%%%%%%%%%%%%
% APPENDIX
%%%%%%%%%%%%%%%%%%%%%%%%%%%%%%%%%%%%%%%%%%%%%%%%%%%%%%%%%%%%%%%%%%%%%%%%%%%%%%%
%%%%%%%%%%%%%%%%%%%%%%%%%%%%%%%%%%%%%%%%%%%%%%%%%%%%%%%%%%%%%%%%%%%%%%%%%%%%%%%
\newpage
\appendix
\onecolumn

\section{Algorithm}
\subsection{Cerebellar Architecture}
Algorithm~\ref{alg:cirl} presents the computational procedure of the cerebellum-inspired feedforward network, highlighting the large expansion in the GrC layer, sparse MF–GrC connectivity, and GrC sparse activation.
\begin{figure*}[ht]
\centering
\begin{minipage}{0.5\textwidth} % 调整宽度
\begin{algorithm}[H]
\caption{Cerebellar Architecture}
\label{alg:cirl}
\begin{algorithmic}
   \STATE {\bfseries Input:} state $x$, network parameters $\theta$
   \STATE {\bfseries Output:} action logits $y$
   
   \STATE $feature \gets \text{Feature Extractor}(x)$
   % \STATE $mf \gets \text{ReLU}(\text{MF}(feature))$ 
   \STATE $mf \gets \text{MF}(feature)$ 
   % \COMMENT{Mossy fiber projection}
   % \STATE $gc \gets \text{GrC}(mf)$
   \STATE $gc \gets \text{GrC}(mf;\, \mathcal{M}_{mf\to gc})$
   % \STATE $gc \gets \text{BatchNorm}(gc)$
   % \STATE $gc \gets \text{ReLU}(\text{clamp}(gc, max=10))$
   % \STATE $normalized \gets gc / \|gc\|_2$
   % \STATE $k \gets \max(1, \lfloor 0.01 \cdot \text{dim}(gc) \rfloor)$
   % \STATE $th \gets \text{TopK}(normalized, k)_{\text{min}}$
   % \STATE $gc\_sparse \gets \text{ReLU}(gc \cdot (normalized \ge th) - 0.1 \cdot \text{mean}(gc))$ 
   \STATE $gc_{\text{sparse}} \gets \text{Sparse Activate}(gc)$
   \STATE $pc\_out \gets \alpha_{pc} \cdot    \text{Purkinje}(gc\_sparse)$
   % \STATE $cn\_excite \gets \text{ReLU}(\text{LayerNorm}(\text{MF\_to\_CN}(feature)))$
  \STATE $cn\_excite \gets \text{MF\_to\_CN}(feature)$
   \STATE $cn \gets \text{clamp}(cn\_excite + pc\_out, min=0)$
   \STATE $y \gets \text{CN output layer}(cn)$
   \STATE \textbf{return} $y$
\end{algorithmic}
\end{algorithm}
\end{minipage}
\end{figure*}

\subsection{Dendritic Modulation Mechanism}
Algorithm~\ref{alg:dendritic-gate} presents the computational flow of the dendritic modulation mechanism.
% Algorithm~\ref{alg:dendritic-gate} delineates the computational procedure of the dendritic modulation mechanism. The algorithm aggregates GrC activations, determines a subset of dendritic branches, computes instantaneous and temporally integrated gains, and modulates the input activations accordingly. 
\begin{figure*}[ht]
\centering
\begin{minipage}{0.5\textwidth} % 调整宽度
\begin{algorithm}[H]
\caption{Dendritic Modulation Mechanism}
\label{alg:dendritic-gate}
\begin{algorithmic}
   \STATE {\bfseries Input:} GrC activations $h_{gc}$, dendritic parameters $\Theta_d$
   \STATE {\bfseries Output:} modulated activations $x_{\text{mod}}$, global\_gain
   \STATE $agg \gets \text{Aggregate}(h_{gc})$
   \STATE $b \gets \text{Select Branches}(agg;\, \Theta_d)$
   \STATE $g_{\text{inst}} \gets \text{Dendritic Gate}(b)$
   \STATE $g \gets \text{Temporal Integrate}(g_{\text{inst}})$
   \STATE $x_{\text{mod}} \gets x \odot \text{Gain}(g)$
   \STATE $global\_gain \gets \text{Global Gain}(g)$
   \STATE \textbf{return} $x_{\text{mod}}, global\_gain$
\end{algorithmic}
\end{algorithm}
\end{minipage}
\end{figure*}

\section{Parameters}
\subsection{Pong game environment parameters}
Table~\ref{tab:pong_env_params_3col} lists the key parameters of the Pong game environment, such as the ball speed, paddle length, frames per second (FPS), and the frame stack size used during training.
% \begin{table*}[htbp]
% \centering
% \caption{Custom Pong environment parameters.}
% \label{tab:pong_env_params_compact}
% \setlength{\tabcolsep}{8pt}
% \renewcommand{\arraystretch}{1.1}
% \begin{tabular}{lclc}
% \toprule
% Parameter & Value & Parameter & Value \\
% \midrule
% Ball X speed & 12 & Paddle width & 10 \\
% Ball Y speed & 8 & Paddle height & 80 \\
% Paddle speed & 5 & FPS & 1920 \\
% Max score & 21 & Action noise & 0.0 \\
% Stack size & 4 &  &  \\
% \bottomrule
% \end{tabular}
% \end{table*}

\begin{table*}[htbp]
\centering
\caption{Custom Pong environment parameters.}
\label{tab:pong_env_params_3col}
\setlength{\tabcolsep}{8pt}
\renewcommand{\arraystretch}{1.1}
\begin{tabular}{l c  l c  l c}
\toprule
Parameter & Value & Parameter & Value & Parameter & Value \\
\midrule
Ball X speed & 12  & Paddle width  & 10  & Ball Y speed & 8 \\
Paddle height & 80 & Paddle speed  & 5   & FPS          & 1920 \\
Max score    & 21  & Action noise  & 0.0 & Stack size   & 4 \\
\bottomrule
\end{tabular}
\end{table*}

\subsection{Training parameters}
Table~\ref{tab:pong_train_params_3col} lists the key training parameters used in our experiments, including the model configuration, learning and exploration settings, and other training hyperparameters. All experiments reported in this paper were performed on NVIDIA RTX 4090 GPUs.
% \begin{table*}[htbp]
% \centering
% \caption{Training parameters.}
% \label{tab:pong_train_params_compact}
% \setlength{\tabcolsep}{8pt}
% \renewcommand{\arraystretch}{1.1}
% \begin{tabular}{lclclc}
% \toprule
% Parameter & Value & Parameter & Value \\
% \midrule
% Model type & CIRL / Baseline & GPU ID & 0 / 1 \\
% Input channels & 4 & Batch size & 64 \\
% Num episodes & 1500 & Gamma & 0.99 \\
% Epsilon start & 1.0 & Epsilon end & 0.01 \\
% Epsilon decay & 200000 & Learning rate & 5e-07 \\
% Memory size & 100000 & Dendritic gate & Enabled / Disabled \\
% Target update freq & 1000 & Save every & 500 \\
% \bottomrule
% \end{tabular}
% \end{table*}

\begin{table*}[!htb]
\centering
\caption{Training parameters.}
\label{tab:pong_train_params_3col}
\setlength{\tabcolsep}{8pt}
\renewcommand{\arraystretch}{1.1}
\begin{tabular}{l c  l c}
\toprule
Parameter & Value & Parameter & Value  \\
\midrule
Model type        & Baseline / CDRL  & GPU ID           & 0 / 1   \\ Input channels      & 4   &Batch size        & 64      \\         
Num episodes     & 1500         & Gamma               & 0.99 \\
Epsilon start     & 1.0             & Epsilon end      & 0.01   \\   Epsilon decay       & 200000  & Learning rate     & 5e-07    \\      
Memory size      & 100000    & Dendritic gate   & Enabled / Disabled \\
Target update freq & 1000           & Save every       & 500    \\   
\bottomrule
\end{tabular}
\end{table*}

\section{Robustness Testing}
% Table~\ref{tab:winrate_wide} presents the robustness win rate results of the baseline model and the CDRL model (bold rows) under different combinations of observation and action noise, showing that the cerebellar model outperforms the baseline.
Table~\ref{tab:winrate_wide} presents the robustness evaluation of the baseline and CDRL models, with bold entries indicating the win rate performance of the CDRL model. Observation noise (Obs) refers to Gaussian noise added to the agent’s perceptual input, with a magnitude ranging from 0.00 to 10.00. While action noise (Act) denotes the probability of stochastic perturbations applied to the agent’s actions, ranging from 0.00 to 0.30. The results show that the CDRL model outperforms the baseline model.

\begin{table*}[htbp]
\centering
\caption{Win rate comparison of the CDRL model and the baseline model under different observation and action noise.}
\label{tab:winrate_wide}
\scriptsize
\setlength{\tabcolsep}{2pt} % 缩小列间距
\renewcommand{\arraystretch}{2.25} % 稍微增加行间距
\resizebox{\textwidth}{!}{%
\begin{tabular}{c|cccccc} % 左侧竖线连续
\toprule
\makecell{Obs / \\ Act} & 0.00 & 0.05 & 0.10 & 0.15 & 0.20 & 0.30 \\
\hline
0.00 & \makecell{0.92$\pm$0.04 \\ \textbf{0.99$\pm$0.01}} & \makecell{0.93$\pm$0.04 \\ \textbf{0.98$\pm$0.01}} & \makecell{0.90$\pm$0.04 \\ \textbf{0.97$\pm$0.02}} & \makecell{0.87$\pm$0.07 \\ \textbf{0.95$\pm$0.03}} & \makecell{0.84$\pm$0.06 \\ \textbf{0.93$\pm$0.02}} & \makecell{0.73$\pm$0.13 \\ \textbf{0.86$\pm$0.04}} \\
\cline{2-7}
1.00 & \makecell{0.88$\pm$0.11 \\ \textbf{0.98$\pm$0.01}} & \makecell{0.83$\pm$0.16 \\ \textbf{0.96$\pm$0.01}} & \makecell{0.84$\pm$0.11 \\ \textbf{0.96$\pm$0.03}} & \makecell{0.80$\pm$0.14 \\ \textbf{0.94$\pm$0.03}} & \makecell{0.79$\pm$0.15 \\ \textbf{0.89$\pm$0.06}} & \makecell{0.62$\pm$0.09 \\ \textbf{0.79$\pm$0.08}} \\
\cline{2-7}
2.00 & \makecell{0.58$\pm$0.31 \\ \textbf{0.84$\pm$0.22}} & \makecell{0.54$\pm$0.29 \\ \textbf{0.81$\pm$0.25}} & \makecell{0.54$\pm$0.31 \\ \textbf{0.79$\pm$0.25}} & \makecell{0.51$\pm$0.29 \\ \textbf{0.76$\pm$0.25}} & \makecell{0.52$\pm$0.23 \\ \textbf{0.72$\pm$0.22}} & \makecell{0.42$\pm$0.21 \\ \textbf{0.62$\pm$0.20}} \\
\cline{2-7}
3.00 & \makecell{0.50$\pm$0.29 \\ \textbf{0.73$\pm$0.26}} & \makecell{0.48$\pm$0.30 \\ \textbf{0.70$\pm$0.27}} & \makecell{0.45$\pm$0.23 \\ \textbf{0.72$\pm$0.24}} & \makecell{0.46$\pm$0.25 \\ \textbf{0.66$\pm$0.25}} & \makecell{0.40$\pm$0.21 \\ \textbf{0.61$\pm$0.23}} & \makecell{0.40$\pm$0.18 \\ \textbf{0.57$\pm$0.21}} \\
\cline{2-7}
5.00 & \makecell{0.30$\pm$0.09 \\ \textbf{0.56$\pm$0.25}} & \makecell{0.33$\pm$0.05 \\ \textbf{0.54$\pm$0.23}} & \makecell{0.30$\pm$0.07 \\ \textbf{0.54$\pm$0.22}} & \makecell{0.28$\pm$0.07 \\ \textbf{0.49$\pm$0.21}} & \makecell{0.32$\pm$0.04 \\ \textbf{0.46$\pm$0.23}} & \makecell{0.29$\pm$0.07 \\ \textbf{0.44$\pm$0.20}} \\
\cline{2-7}
10.00 & \makecell{0.27$\pm$0.07 \\ \textbf{0.37$\pm$0.22}} & \makecell{0.31$\pm$0.06 \\ \textbf{0.38$\pm$0.18}} & \makecell{0.29$\pm$0.06 \\ \textbf{0.35$\pm$0.19}} & \makecell{0.26$\pm$0.06 \\ \textbf{0.34$\pm$0.17}} & \makecell{0.28$\pm$0.04 \\ \textbf{0.31$\pm$0.15}} & \makecell{0.27$\pm$0.01 \\ \textbf{0.28$\pm$0.14}} \\
\bottomrule
\end{tabular}}
\end{table*}

% Table~\ref{tab:winrate_cirl} presents the robustness evaluation of the baseline model and CDRL-dendrite model (bold rows) under different combinations of observation and action noise, showing that the cerebellar model outperforms the baseline in most cases.
Table~\ref{tab:winrate_cirl} presents the robustness evaluation of the baseline and CDRL-dendrite models, with bold rows indicating the win rate of the CDRL-dendrite model under different combinations of observation and action noise. The results show that the CDRL-dendrite model consistently achieves higher win rate than the baseline across most test conditions.

\begin{table*}[htbp]
\centering
\caption{Win rate comparison of the CDRL-dendrite model and the baseline model under different observation and action noise.}
\label{tab:winrate_cirl}
\scriptsize
\setlength{\tabcolsep}{2pt}
\renewcommand{\arraystretch}{2.25}
\resizebox{\textwidth}{!}{%
\begin{tabular}{c|cccccc}
\toprule
\makecell{Obs / \\ Act} & 0.00 & 0.05 & 0.10 & 0.15 & 0.20 & 0.30 \\
\hline
0.00 &
\makecell{0.92$\pm$0.04 \\ \textbf{1 $\pm$0.01}} &
\makecell{0.93$\pm$0.04 \\ \textbf{0.97$\pm$0.01}} &
\makecell{0.90$\pm$0.04 \\ \textbf{0.96$\pm$0.03}} &
\makecell{0.87$\pm$0.07 \\ \textbf{0.97$\pm$0.01}} &
\makecell{0.84$\pm$0.06 \\ \textbf{0.95$\pm$0.03}} &
\makecell{0.73$\pm$0.13 \\ \textbf{0.92$\pm$0.05}} \\
\cline{2-7}
1.00 &
\makecell{0.88$\pm$0.11 \\ \textbf{0.90$\pm$0.10}} &
\makecell{0.83$\pm$0.16 \\ \textbf{0.92$\pm$0.08}} &
\makecell{0.84$\pm$0.11 \\ \textbf{0.90$\pm$0.09}} &
\makecell{0.80$\pm$0.14 \\ \textbf{0.91$\pm$0.08}} &
\makecell{0.79$\pm$0.15 \\ \textbf{0.88$\pm$0.09}} &
\makecell{0.62$\pm$0.09 \\ \textbf{0.85$\pm$0.09}} \\
\cline{2-7}
2.00 &
\makecell{0.58$\pm$0.31 \\ \textbf{0.62$\pm$0.24}} &
\makecell{0.54$\pm$0.29 \\ \textbf{0.64$\pm$0.23}} &
\makecell{0.54$\pm$0.31 \\ \textbf{0.63$\pm$0.27}} &
\makecell{0.51$\pm$0.29 \\ \textbf{0.62$\pm$0.27}} &
\makecell{0.52$\pm$0.23 \\ \textbf{0.59$\pm$0.24}} &
\makecell{0.42$\pm$0.21 \\ \textbf{0.60$\pm$0.21}} \\
\cline{2-7}
3.00 &
\makecell{0.50$\pm$0.29 \\ \textbf{0.48$\pm$0.23}} &
\makecell{0.48$\pm$0.30 \\ \textbf{0.51$\pm$0.26}} &
\makecell{0.45$\pm$0.23 \\ \textbf{0.46$\pm$0.26}} &
\makecell{0.46$\pm$0.25 \\ \textbf{0.51$\pm$0.21}} &
\makecell{0.40$\pm$0.21 \\ \textbf{0.46$\pm$0.24}} &
\makecell{0.40$\pm$0.18 \\ \textbf{0.49$\pm$0.20}} \\
\cline{2-7}
5.00 &
\makecell{0.30$\pm$0.09 \\ \textbf{0.34$\pm$0.24}} &
\makecell{0.33$\pm$0.05 \\ \textbf{0.31$\pm$0.23}} &
\makecell{0.30$\pm$0.07 \\ \textbf{0.33$\pm$0.24}} &
\makecell{0.28$\pm$0.07 \\ \textbf{0.30$\pm$0.18}} &
\makecell{0.32$\pm$0.04 \\ \textbf{0.31$\pm$0.18}} &
\makecell{0.29$\pm$0.07 \\ \textbf{0.32$\pm$0.15}} \\
\cline{2-7}
10.00 &
\makecell{0.27$\pm$0.07 \\ \textbf{0.29$\pm$0.05}} &
\makecell{0.31$\pm$0.06 \\ \textbf{0.28$\pm$0.04}} &
\makecell{0.29$\pm$0.06 \\ \textbf{0.31$\pm$0.07}} &
\makecell{0.26$\pm$0.06 \\ \textbf{0.31$\pm$0.08}} &
\makecell{0.28$\pm$0.04 \\ \textbf{0.28$\pm$0.06}} &
\makecell{0.27$\pm$0.01 \\ \textbf{0.27$\pm$0.06}} \\
\bottomrule
\end{tabular}}
\end{table*}

\section{Code Availability}
To facilitate reproducibility, the code and experimental setups for the CDRL framework will be made publicly available following publication.

\end{document}